\documentclass[pdflatex,sn-mathphys-num]{sn-jnl}

\usepackage{graphicx}%
\usepackage{multirow}%
\usepackage{amsmath,amssymb,amsfonts}%
\usepackage{amsthm}%
\usepackage{mathrsfs}%
\usepackage[title]{appendix}%
\usepackage{xcolor}%
\usepackage{textcomp}%
\usepackage{manyfoot}%
\usepackage{booktabs}%
\usepackage{algorithm}%
\usepackage{algorithmicx}%
\usepackage{algpseudocode}%
\usepackage{listings}%
\usepackage{url}%
\usepackage{tabularray}

\theoremstyle{thmstyleone}%

\theoremstyle{thmstyletwo}%

\theoremstyle{thmstylethree}%

\begin{document}

\title[Core-Boundary Awareness for Resampling]{Oversampling and Downsampling with Core-Boundary Awareness: A Data Quality-Driven Approach}

\author[1]{\fnm{Samir Brahim} \sur{Belhaouari}}\email{sbelhaouari@hbku.edu.qa}

\author[1]{\fnm{Yunis Carreon} \sur{Kahalan}}\email{yuka34154@hbku.edu.qa}

\author[1]{\fnm{Humaira} \sur{Shaffique}}\email{hush34114@hbku.edu.qa}

\author[2]{\fnm{Ismael} \sur{Belhaouari}}\email{i.brahimbelhaouari@student.maastrichtuniversity.nl}

\author[3]{\fnm{Ashhadul} \sur{Islam}}\email{aisla@kth.se}

\affil[1]{\orgname{Hamad Bin Khalifa University}, \orgaddress{\city{Doha}, \country{Qatar}}}

\affil[2]{\orgname{Maastricht University}, \orgaddress{\city{Maastricht}, \country{Netherlands}}}

\affil[3]{\orgname{KTH Royal Institute of Technology}, \orgaddress{\city{Stockholm}, \country{Sweden}}}

\abstract{
The effectiveness of machine learning models, particularly in unbalanced classification tasks, is often hindered by the failure to differentiate between critical instances near the decision boundary and redundant samples concentrated in the core of the data distribution. In this paper, we propose a method to systematically identify and differentiate between these two types of data. Through extensive experiments on multiple benchmark datasets, we show that the boundary data oversampling method improves the F1 score by up to 10\% on 96\% of the datasets, whereas our core-aware reduction method compresses datasets up to 90\% while preserving their accuracy, making it 10 times more powerful than the original dataset. Beyond imbalanced classification, our method has broader implications for efficient model training, particularly in computationally expensive domains such as Large Language Model (LLM) training. By prioritizing high-quality, decision-relevant data, our approach can be extended to text, multimodal, and self-supervised learning scenarios, offering a pathway to faster convergence, improved generalization, and significant computational savings. This work paves the way for future research in data-efficient learning, where intelligent sampling replaces brute-force expansion, driving the next generation of AI advancements. Our code is available as a Python package at \url{https://pypi.org/project/adaptive-resampling/}.
}

\keywords{Classification, Core-Boundary Awareness, Decision Boundary, Oversampling, Downsampling, Large Language Models (LLM), Data Quality}

\maketitle

\section{Introduction}
 In data science and machine learning, class imbalance is a significant issue, particularly in applications such as medical diagnosis where a class can be rare. The cost of misclassifying minority instances is high. When training on imbalanced data sets, models tend to focus on the majority class, resulting in biased predictions and low accuracy for minority class instances \cite{weiss2004mining}. Oversampling and downsampling are two primary techniques used to mitigate these biases by modifying the distribution of data and thus enhancing model performance in class-imbalanced settings.

Oversampling methods increase the representation of the minority class, either by replicating existing samples or by generating synthetic samples. A basic approach is random oversampling, which duplicates minority class samples to balance the class distribution. However, this can lead to overfitting because the model may learn repeated patterns in the duplicated samples reducing generalizability. A more advanced method is SMOTE (Synthetic Minority Over-sampling Technique), introduced by Chawla et al \cite{chawla2002smote}. SMOTE generates synthetic samples by interpolating between existing minority class instances, enhancing minority representation without simple duplication. ADASYN (Adaptive Synthetic Sampling) further refines this by generating synthetic samples for harder-to-classify minority instances enhancing classification accuracy for difficult samples \cite{he2008adasyn}.

Downsampling methods, by contrast, reduce the majority class representation. A straightforward approach is random downsampling, which balances the dataset by randomly discarding majority samples. While effective, random downsampling risks losing valuable information and potentially leads to underfitting \cite{lemaitre2017imbalanced}. Cluster-based downsampling approaches, such as MWMOTE, use clustering algorithms to strategically select and reduce majority samples, in order to retain the diversity of the majority class while balancing the dataset \cite{barua2014mwmote}.

The rest of this paper is organized as follows. Section~\ref{sec:RelatedWork} reviews related work on oversampling and downsampling on imbalanced data sets. Section~\ref{sec:Methodology} details the proposed methodology on the core and boundary aware approach. Section~\ref{sec:data_experiment} describes the datasets and the experimental setups using the core and boundary aware approach. Section~\ref{sec:Result} presents and discusses the results for each experiment. Section~\ref{sec:Conc} concludes the paper and outlines the future research direction. Finally, Section~\ref{sec:appendix} gathers supplementary material in the Appendix.

\section{Related Work}\label{sec:RelatedWork}

In the realm of imbalanced data classification, several methodologies have been developed to identify and leverage core and border points within datasets. A prominent technique is Borderline-SMOTE \cite{han2005borderline}, an enhancement of the Synthetic Minority Over-sampling Technique (SMOTE). This approach focuses on generating synthetic samples specifically for minority class instances that are situated near the decision boundary, or on the border between classes. By concentrating on these critical borderline instances, Borderline-SMOTE aims to bolster the classifier’s ability to discern between classes, thereby improving predictive performance. 

Another significant method is the Density-Based Spatial Clustering of Applications with Noise (DBSCAN) \cite{ester1996density} algorithm. DBSCAN is adept at clustering data by identifying core points, which are data points encompassed by a sufficient number of neighboring points within a specified radius. Points that do not meet this criterion but are within the neighborhood of core points are classified as border points, while those that are neither core nor border points are considered noise. This clustering technique effectively distinguishes between densely populated regions and sparser areas, facilitating the identification of core and border points within the data distribution. 

Furthermore, the DBSM approach \cite{sanguanmak2016dbsm}, which combines DBSCAN and SMOTE, has been proposed to address imbalanced data classification. This method utilizes DBSCAN to identify core and border points and then applies SMOTE to generate synthetic samples, enhancing the representation of the minority class.

\subsection{Other Oversampling Algorithms}

Beyond the identification of core and border points, various state-of-the-art oversampling techniques have been proposed to address class imbalance by generating synthetic samples. These methods often extend or refine SMOTE to improve data representation and classification performance.

\textbf{SMOTE-ENN} \cite{batista2004study} is a hybrid technique that combines SMOTE with Edited Nearest Neighbors (ENN). While SMOTE generates synthetic samples to balance the dataset, ENN removes mislabeled or noisy instances, leading to a more refined training set. \textbf{Polynomial Fit SMOTE} (Bus, Mesh, Poly, Star) \cite{Gazzah2008} is a family of oversampling techniques that utilize polynomial regression to interpolate new synthetic samples. These methods focus on better approximating the underlying data distribution compared to traditional SMOTE. \textbf{ProWSyn} (Progressive Weighted Synthetic Oversampling) \cite{Barua2013} is an adaptive oversampling method that assigns different weights to synthetic samples based on their proximity to the decision boundary. This dynamic weighting mechanism ensures that new samples are more informative. \textbf{SMOTE-IPF} \cite{Saez2015a} integrates the Iterative Partitioning Filter (IPF) with SMOTE to iteratively refine the synthetic sample set by filtering out potentially harmful instances, thereby improving classification robustness. \textbf{Lee’s Method} \cite{Lee2015} enhances oversampling by employing a neighborhood-based approach that adjusts synthetic sample placement to better reflect the distribution of minority class instances. \textbf{SMOBD} \cite{Cao2011} (SMOTE with Border Detection) explicitly identifies and prioritizes borderline instances for oversampling, reinforcing decision boundary learning. \textbf{G-SMOTE} (Geometric SMOTE) \cite{Sandhan2014} extends SMOTE by leveraging geometric transformations, such as rotations and scaling, to generate synthetic samples with improved diversity. \textbf{CCR} (Cluster-based Classifier-based Resampling) \cite{Koziarski2017} applies clustering to minority class instances before oversampling, ensuring that synthetic samples are generated within meaningful subgroups. \textbf{Assembled SMOTE} \cite{Zhou2013} integrates multiple oversampling techniques to provide a comprehensive minority class augmentation strategy. \textbf{SMOTE-Tomek Links} \cite{Batista2004} refines SMOTE by applying Tomek links, a method that removes samples contributing to class overlap, thereby improving class separability. We also take a look at some of the downsampling algorithms that are used.

\subsection{Other Undersampling Algorithms} 
In addition to oversampling techniques, various downsampling methods have been developed to address class imbalance by reducing the number of majority class instances. These techniques aim to retain critical samples while eliminating redundant or potentially misleading data points, ensuring that the classifier focuses on the most informative regions of the data distribution. \textbf{Neighbourhood Cleaning Rule (NCR)} \cite{laurikkala2001improving} refines the dataset by removing majority class instances that are likely to be misclassified by their nearest neighbors. By iteratively cleaning the dataset, NCR enhances decision boundary clarity while preserving representative samples. \textbf{One-Sided Selection (OSS)} \cite{kubat1997addressing} combines Tomek Links and Condensed Nearest Neighbors (CNN) to selectively remove redundant majority instances while retaining critical samples that contribute to class separability. This approach is particularly effective for maintaining an informative dataset. \textbf{Condensed Nearest Neighbour (CNN)} \cite{hart1968condensed} is a prototype selection method that iteratively removes redundant majority class points, ensuring that only the essential samples are retained to define class boundaries. \textbf{AllKNN} \cite{tomek1976experiment} extends the Edited Nearest Neighbors (ENN) technique by iteratively removing samples that introduce noise in nearest-neighbor-based classification. By reducing class overlap, this method improves classifier generalization. \textbf{Edited Nearest Neighbours (ENN)} \cite{wilson1972asymptotic} refines the dataset by removing instances that are misclassified by their k-nearest neighbors. This process improves class distinction by eliminating noisy majority class samples. \textbf{Cluster Centroids (CC)} \cite{yen2009cluster} applies k-means clustering to the majority class and replaces original samples with the cluster centroids. This technique reduces redundancy while preserving the overall class distribution. \textbf{Random UnderSampler (RUS)} \cite{drummond2003c4} randomly removes majority class instances to achieve class balance. Although it is computationally efficient, this approach may discard informative samples, potentially weakening the classifier’s decision boundaries. These downsampling methods offer distinct strategies for reducing the dataset size while maintaining classification performance. When applied in conjunction with oversampling techniques, they can create a more balanced dataset, leading to improved model generalization and robustness in handling class imbalance.

Our contributions can be summarised as follows:

\begin{itemize}
    \item We introduce a novel borderline-aware oversampling method that selectively augments critical minority class instances located near decision boundaries. This targeted augmentation strategy improves class separability and enhances classifier generalization, achieving an average F1 score improvement of 4.26\% across multiple benchmark datasets.
    \item We propose a core-aware data reduction strategy that systematically prunes redundant majority class instances while preserving key decision boundaries. Our experimental results demonstrate that up to 25\% of the dataset can be removed without any performance degradation, thereby optimizing computational efficiency without sacrificing classification accuracy.
    \item Our approach highlights the importance of data quality over raw data volume, paving the way for efficient training of Large Language Models (LLMs) and other resource-intensive AI systems. By extending our method to prioritize high-quality, decision-relevant data in large-scale learning scenarios, we provide a promising direction for reducing the computational cost of training AI models while improving convergence and generalization.
\end{itemize}

\section{Methodology}\label{sec:Methodology}

\subsection{Conceptual Framework}

In this part, we define key terms relevant to our approach, including majority class, minority class, core points, and border points, which are crucial for understanding the proposed oversampling method.

\textbf{Majority Class:} The class with a higher number of instances in an imbalanced dataset. Traditional classifiers tend to be biased towards this class due to its dominant representation.

\textbf{Minority Class:} The underrepresented class in an imbalanced dataset. Due to its lower frequency, classifiers often struggle to learn effective decision boundaries for this class.

\textbf{Core Points:} Core points are data samples that lie deep within the majority or minority class clusters. These points are well-represented within their respective classes and do not contribute significantly to defining the decision boundary.

\textbf{Border Points:} Border points are those that exist near the decision boundary between the majority and minority classes. These points are critical for classification, as they define the separation between the two classes.

\begin{figure}[htbp]
    \centering
    \includegraphics[width=0.5\textwidth]{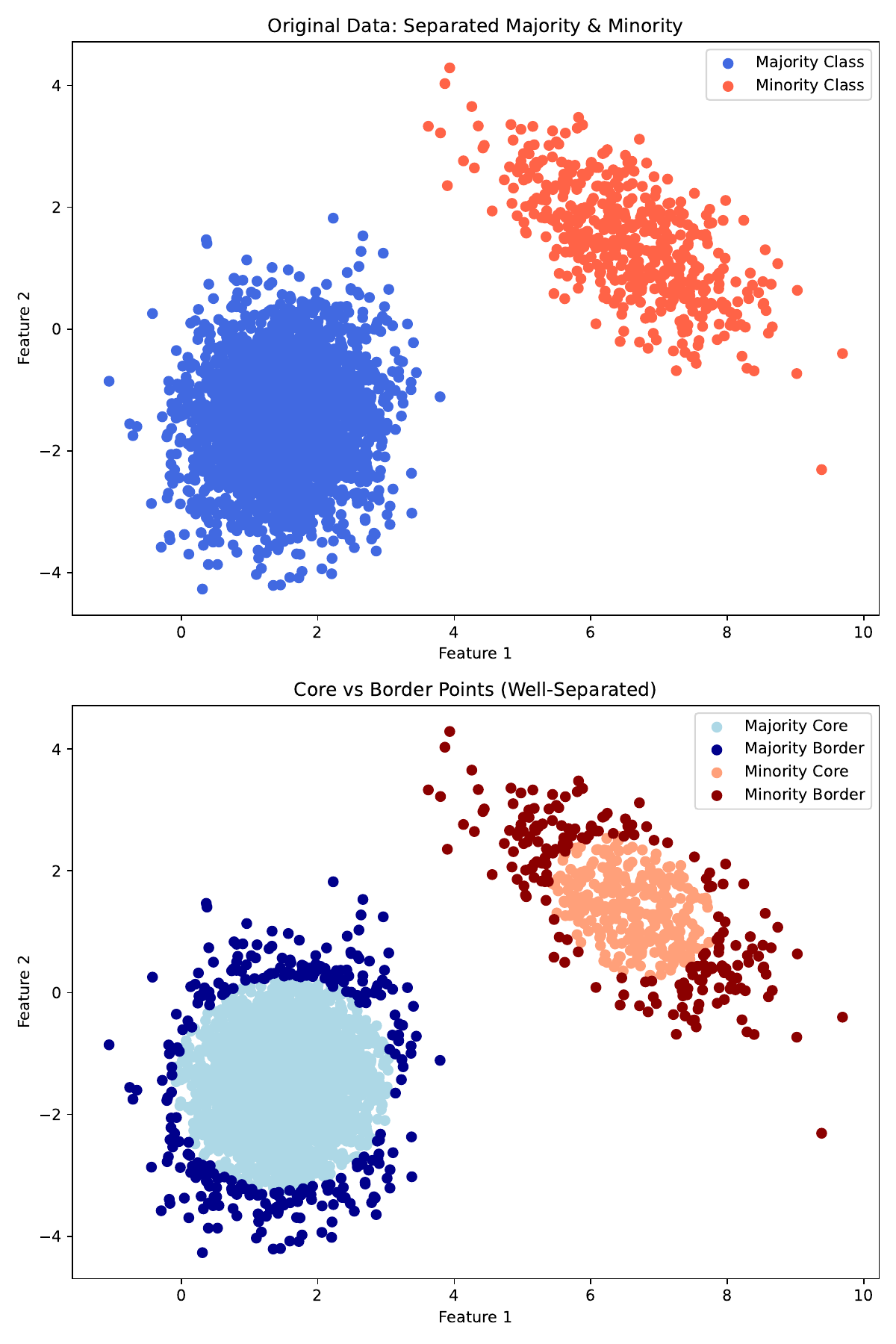}
    \caption{Visualization of the Separation Between Majority and Minority Classes and the Identification of Core and Border Points. The top plot shows the original dataset with well-separated majority and minority classes, while the bottom plot highlights the core and border points within each class.}
    \label{fig:border_core_trivial}
\end{figure}

Figure~\ref{fig:border_core_trivial} illustrates these concepts in a classification setting. The top plot represents a trivial case where the majority class (blue) and minority class (red) are well-separated, providing a clear distinction between the two. The bottom plot further refines this representation by distinguishing core points (lighter shades) and border points (darker shades) for both classes. Core points are embedded deep within their respective clusters, whereas border points are positioned closer to the decision boundary. This distinction is crucial for oversampling strategies, as augmenting border points rather than core points ensures that new synthetic samples better reinforce the decision boundary, thereby improving classification performance in imbalanced datasets.

\subsection{Identifying Core and Border Points}

\begin{figure}[htbp]
    \centering
    \includegraphics[width=0.4\textwidth]{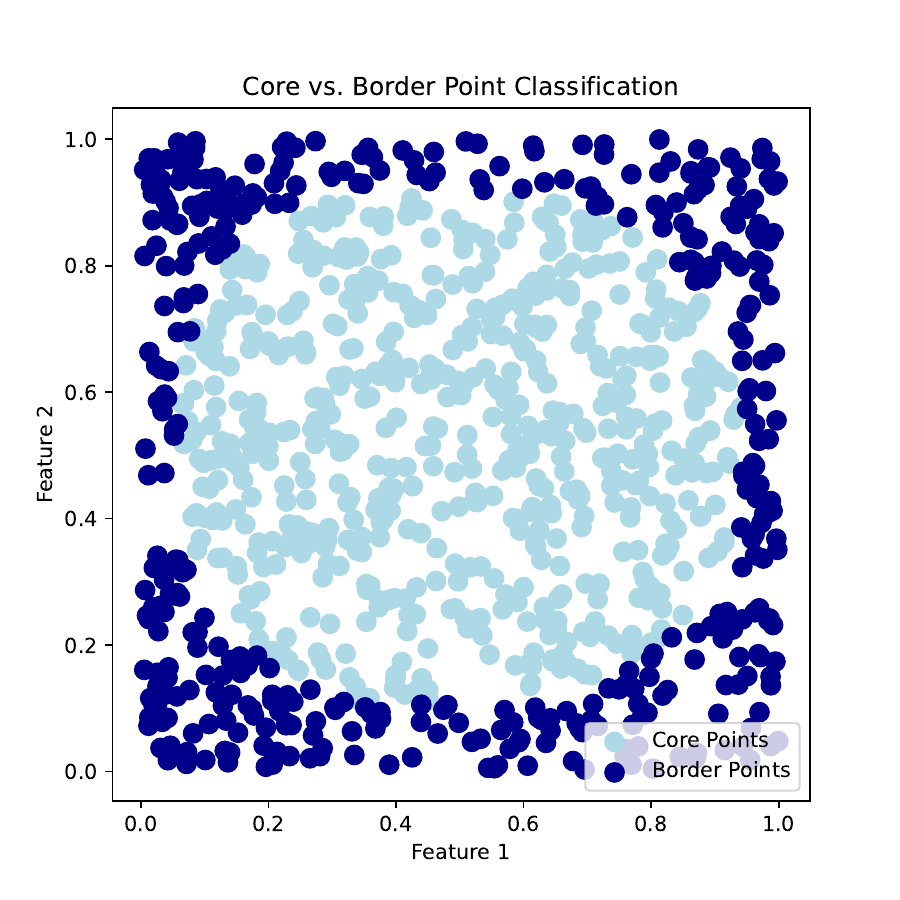}
    \caption{Illustration of Core and Border Points in a Two-Dimensional Feature Space. Core points lie deep within clusters, while border points are near the decision boundary—useful in selective oversampling strategies to improve classifier performance.}
    \label{fig:simple_borderVcore}
\end{figure}

To enhance the performance of classification models in imbalanced datasets, we introduce a method for identifying core and border points in a dataset. This method utilizes a distance-based approach to classify points based on their relative proximity to neighboring samples as seen in Figure~\ref{fig:simple_borderVcore}.

\subsubsection{Distance-Based Classification}

Given a dataset \( X = \{x_1, x_2, ..., x_n\} \), where each \( x_i \in \mathbb{R}^d \) represents a feature vector in a \( d \)-dimensional space, we define the distance metric between any two points \( x_i \) and \( x_j \) using the \( p \)-norm:

\[
d_p(x_i, x_j) = \left( \sum_{k=1}^{d} |x_{i,k} - x_{j,k}|^p \right)^{\frac{1}{p}}
\]

where \( p=2 \) corresponds to the Euclidean distance, while other values of \( p \) can be chosen for different distance metrics. We define the average distance of a point \( x_i \) from its \( k \)-nearest neighbors as:

\[
d_{x_i} = \frac{1}{k} \sum_{j=1}^{k} d\left(x_i, x_j^k \right)
\]

where \( x_j^k \) represents the \( j^{th} \) nearest neighbor of \( x_i \), and \( d(x_i, x_j^k) \) denotes the distance metric (e.g., Euclidean distance when \( p=2 \)) between \( x_i \) and \( x_j^k \).

To determine the classification threshold for identifying core and border points, we compute a distance threshold \( d_t \) based on the percentile \( P_{\alpha} \) of the distance distribution among all points in a given class:

\[
d_t = P_{\alpha} \left\{ d_{x_i} \mid x_i \in \text{class} \right\}
\]

\subsubsection{Classification of Core and Border Points}

- A point \( x_i \) is classified as a \textbf{border point} if its average distance to its \( k \)-nearest neighbors exceeds the threshold \( d_t \):

  \[
  x_i \in \text{Border} \quad \text{if} \quad d_{x_i} > d_t
  \]

- A point \( x_i \) is classified as a \textbf{core point} if its average distance is smaller than or equal to \( d_t \):

  \[
  d_{x_i} \leq d_t \Rightarrow x_i \in \text{Core}
  \]

\[
\text{where } B = \text{border}, \quad C = \text{cluster}
\]

This classification method ensures that core points are those embedded within dense regions of the dataset, while border points are located near the decision boundary. This distinction is crucial for oversampling strategies, as augmenting border points rather than core points ensures that new synthetic samples reinforce the decision boundary, thereby improving classification performance in imbalanced datasets.




However, in some datasets, points may lie very close to each other, making it difficult to distinguish between core and boundary points, as shown in Figure~\ref{fig:donut} below. This could lead to problems in oversampling and downsampling which in turn could affect model training and results. One common consequence is overfitting, where the model memorizes minor local variations caused by densely packed points, resulting in poor generalization to new data.

\begin{figure}[htbp]
    \centering
    \includegraphics[width=0.5\textwidth]{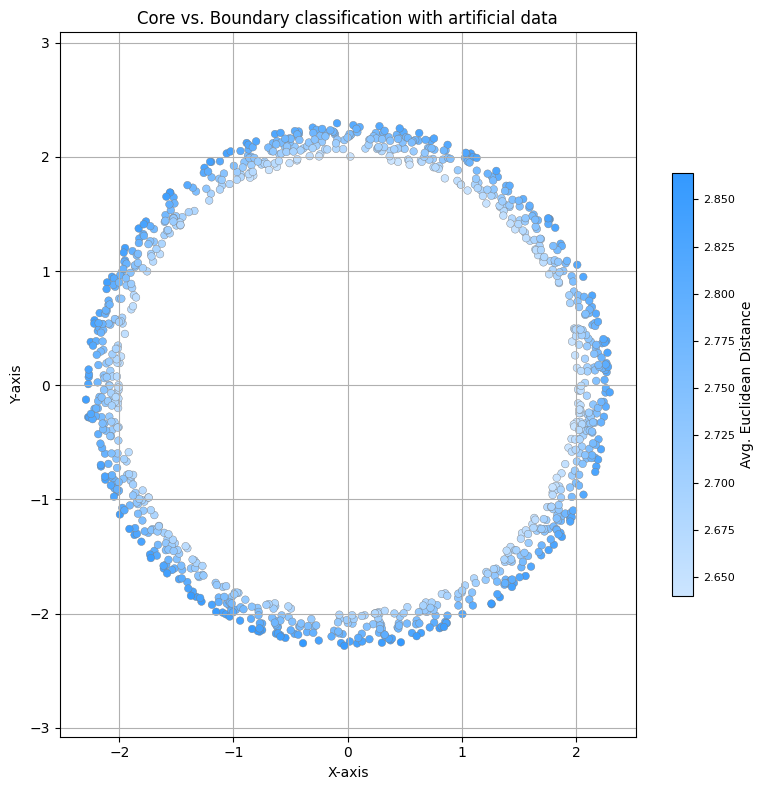}
    \caption{Visualization of Core Versus Boundary Classification in a Two-Dimensional Donut-Shaped Dataset. Due to uniform density and proximity between points, core and boundary classifications become ambiguous.}
    \label{fig:donut}
\end{figure}

\subsection{Balanced Sampling Strategies}

\begin{figure}[htbp]
    \centering
    \includegraphics[width=0.5\textwidth]{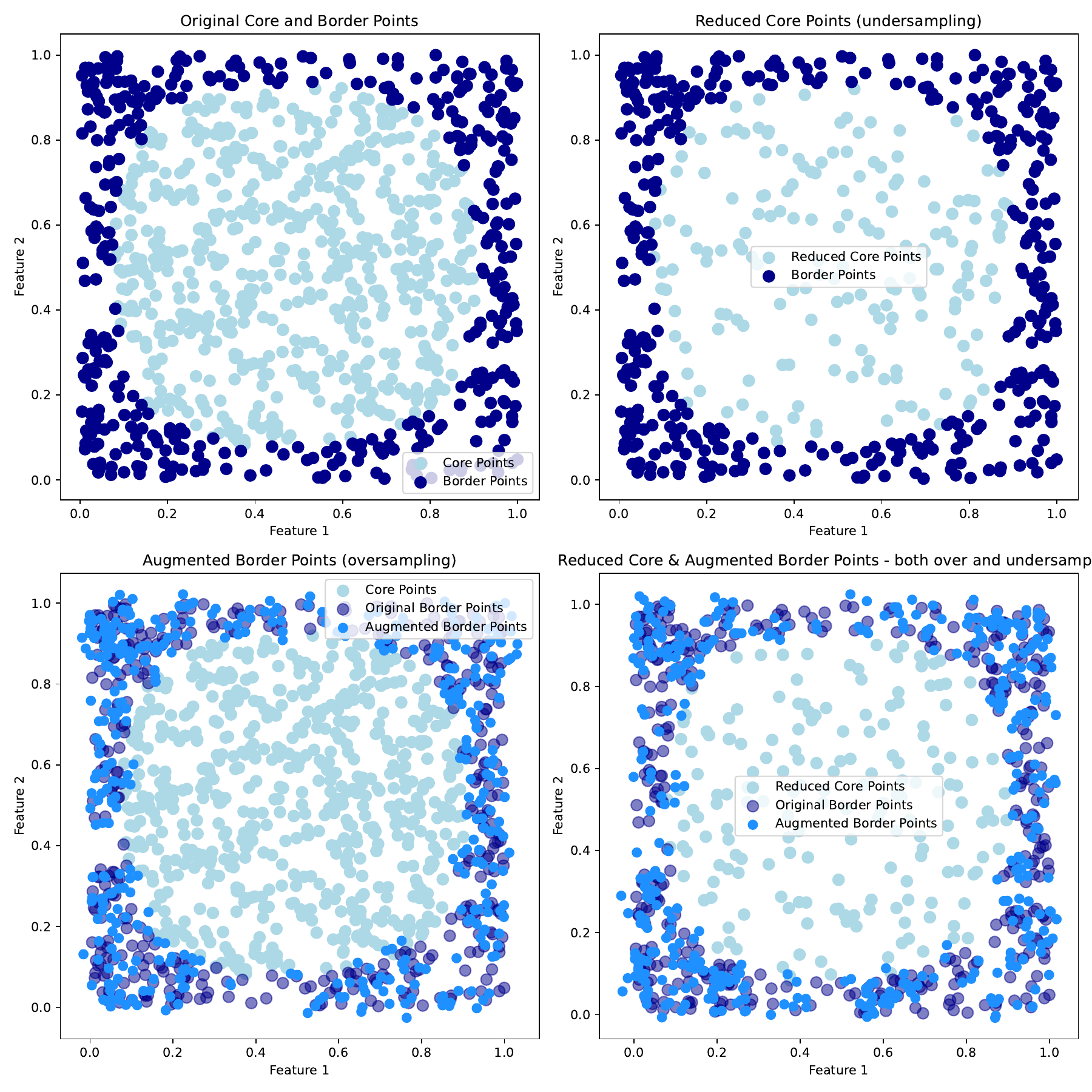}
     \caption{Illustration of Different Sampling Strategies Applied to Core and Border Points. The subplots show: (1) the original distribution, (2) downsampling by reducing core points, (3) oversampling by augmenting border points, and (4) a combination of both strategies.}
    \label{fig:over_under_sampling}
\end{figure}

Figure~\ref{fig:over_under_sampling} illustrates different strategies for handling imbalanced datasets after identifying core and border points. The first subplot represents the original dataset where core and border points are clearly distinguished. The second subplot shows downsampling, where a proportion of core points are removed while preserving the border points. This strategy effectively reduces data redundancy, ensuring that the classifier focuses more on informative decision boundaries. The third subplot demonstrates oversampling, where a subset of border points is selected and augmented using synthetic variations, thereby reinforcing minority class representation. The fourth subplot combines both downsampling and oversampling, reducing core points while augmenting border points. This hybrid approach balances the dataset efficiently by maintaining class boundaries while preventing overfitting caused by excessive duplication of core samples. These techniques, when applied prudently, enhance classification robustness in imbalanced datasets by improving the generalization of decision boundaries without artificially distorting data distributions.

\section{Data And Experiment}\label{sec:data_experiment}
To test the core and boundary aware approach, two different experiment setups are built. The first experiment oversamples the minority border points along with additionally downsampling the majority core points. The second experiment focuses on only downsampling the majority core points. 

\subsection{Experiment I : Oversampling borderline data}

\begin{figure*}[htbp]
    \centering
    \includegraphics[width=\textwidth]{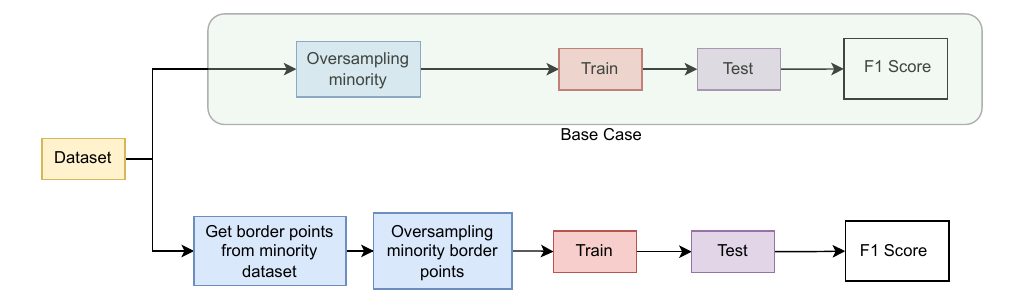} 
    \caption{Workflow of the Experimentation Methodology. The base case involves oversampling the entire minority dataset, while the proposed method focuses on identifying and augmenting only the borderline points from the minority dataset before training the model. Key performance metrics such as F1 score, accuracy, precision, and recall are compared between the two approaches.}
    \label{fig:experiment_workflow}
\end{figure*}

\begin{figure*}[htbp]
\centering
\includegraphics[width=\textwidth]{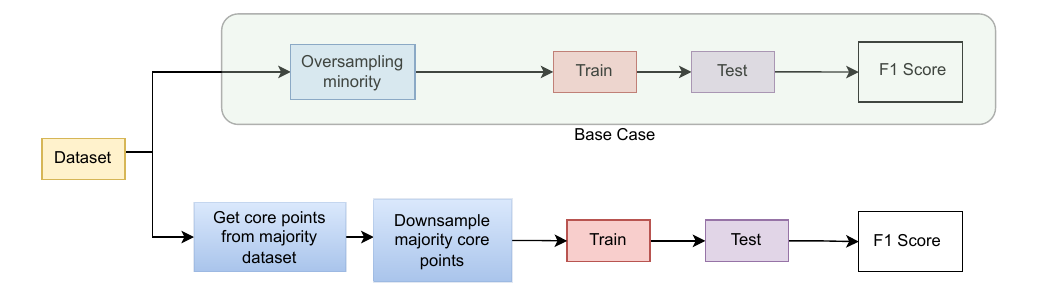}
\caption{Illustration of the Data Processing Pipeline Incorporating Both Oversampling and Downsampling Strategies. The baseline approach involves traditional oversampling of the entire minority dataset and evaluating performance based on the F1 score. The proposed method refines this by identifying core points within the majority class and selectively downsampling them.}
\label{fig:Borderline_Experiment_downsample}
\end{figure*}

Figure~\ref{fig:experiment_workflow} presents the first stage of our methodology, where the baseline approach is compared against our proposed method of targeted borderline oversampling. This figure highlights the impact of oversampling only the minority class instances that are located near the decision boundary, rather than oversampling the entire minority dataset indiscriminately. In contrast, Figure~\ref{fig:Borderline_Experiment_downsample} extends this methodology by incorporating a downsampling strategy. The baseline approach remains present in both figures, serving as a control. However, our full approach goes beyond oversampling the minority class by also reducing redundancy in the majority class. We achieve this by detecting core points within the majority class—those that are far from the decision boundary and do not contribute significantly to class separation—and strategically removing them. By combining borderline oversampling and core-aware downsampling, we create a more balanced and information-dense dataset that enhances classifier learning.

To evaluate the effectiveness of our proposed method, we conducted a comprehensive series of experiments on a diverse set of datasets. The details of each dataset is given in the appendix in Table \ref{tab:dataset_characteristics}

\subsubsection{Oversampling Algorithms}

The oversampling algorithms considered in the experiments are listed below: 

\begin{itemize}
    \item SMOTE-ENN \cite{batista2004study}
    \item Polynomial Fit SMOTE (Bus) \cite{Gazzah2008}
    \item Polynomial Fit SMOTE (Mesh) \cite{Gazzah2008}
    \item Polynomial Fit SMOTE (Poly) \cite{Gazzah2008}
    \item Polynomial Fit SMOTE (Star) \cite{Gazzah2008}
    \item ProWSyn \cite{Barua2013}
    \item SMOTE-IPF \cite{Saez2015a}
    \item Lee \cite{Lee2015}
    \item SMOBD \cite{Cao2011}
    \item G-SMOTE \cite{Sandhan2014}
    \item CCR \cite{Koziarski2017}
    \item Assembled SMOTE \cite{Zhou2013}
    \item SMOTE-Tomek Links \cite{Batista2004}
    \item No Oversampling (NoSMOTE) (Special case for no oversampling)
\end{itemize}

\subsubsection{Classifiers}

To ensure comprehensive evaluation, we employed a diverse set of machine learning classifiers. The classifiers used are as follows:
\begin{itemize}
    \item Random Forest Classifier \cite{breiman2001random}
    \item Multi-Layer Perceptron (MLP) Classifier \cite{hinton1990connectionist}
    \item Support Vector Machine (SVM) \cite{platt2007probabilistic}
    \item Decision Tree Classifier \cite{dumont2009fast}
    \item AdaBoost Classifier \cite{hastie2009multi}
    \item XGBoost Classifier \cite{chen2016xgboost}
    \item CatBoost Classifier \cite{prokhorenkova2018catboost}
\end{itemize}

\subsection{Experiment II: Downsampling core datapoints}

In this phase of the experiment, we investigate the impact of strategic data reduction on large datasets that contain a large number of samples. Our approach leverages borderline and core detection to selectively downsample data points while preserving critical decision boundaries. Specifically, we analyze the majority class within each dataset, identifying both the core and boundary points. The goal is to prune non-essential core samples while retaining the borderline instances, ensuring that the model continues to learn meaningful class distinctions. After applying this reduction, we train the classifiers on the refined dataset and evaluate their performance on the test data. We hypothesize that this targeted reduction maintains model effectiveness as decision boundaries remain well represented, avoiding significant performance degradation whilst optimizing dataset efficiency.

\begin{figure}[htbp]
    \centering
    \includegraphics[width=0.8\textwidth]{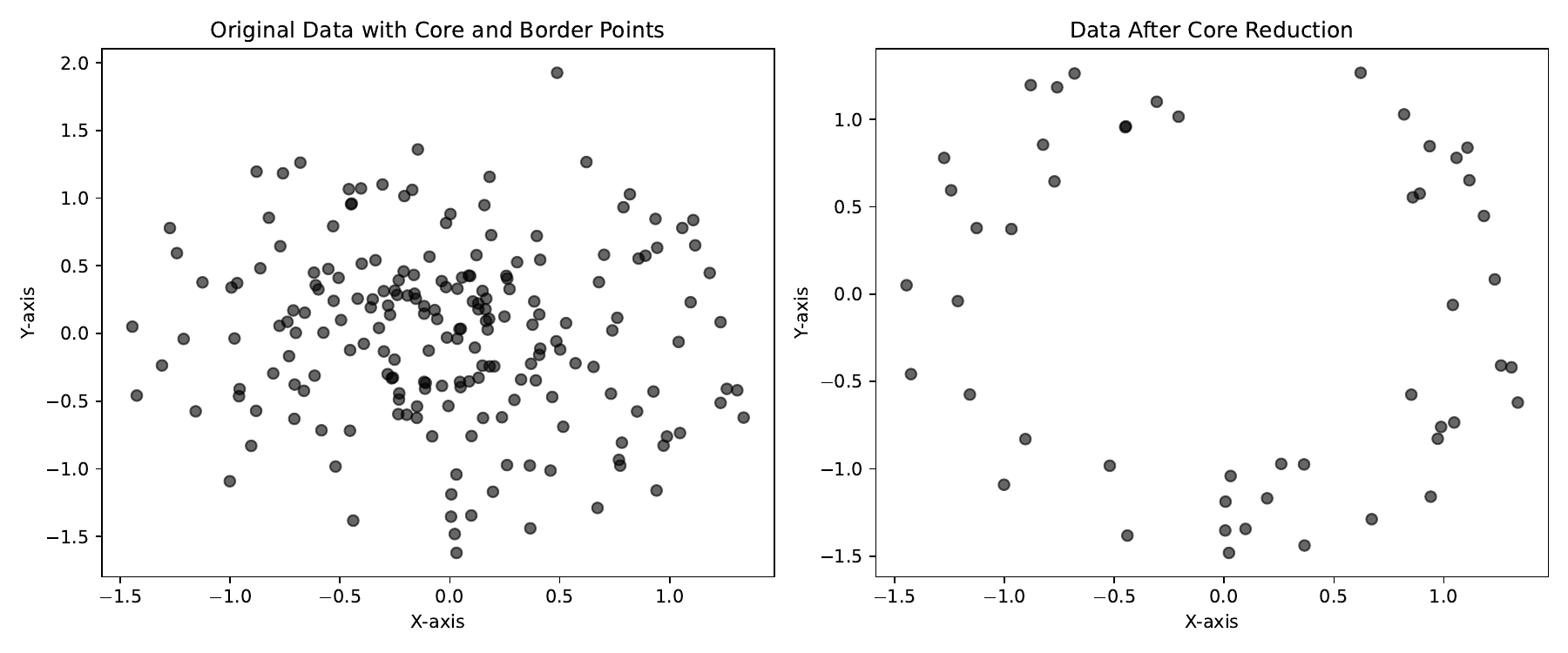}
    \caption{Visualization of Data Reduction Using Core and Borderline Detection.}
    \label{fig:core_removal}
\end{figure}

Figure \ref{fig:core_removal} shows the method - left plot illustrates the original dataset containing both core and border points, while the right plot shows the dataset after core reduction, where only the critical borderline points are retained. This selective pruning ensures that decision boundaries remain intact while reducing redundancy in the dataset.

\subsubsection{Datasets}
For this part of the experiment, we use four datasets, each representing a distinct domain:

\begin{itemize}
    \item \textbf{Air Quality Dataset} \cite{air_quality_360}:
    This dataset contains atmospheric pollution measurements collected from an air quality monitoring station. It includes various sensor readings such as CO(GT) (carbon monoxide concentration), NMHC(GT) (non-methane hydrocarbons), C6H6(GT) (benzene concentration), NOx(GT) (nitrogen oxides), NO2(GT) (nitrogen dioxide), O3 (ozone), temperature (T), relative humidity (RH), and absolute humidity (AH). Each record represents hourly measurements of these pollutants and meteorological factors. For our experiment, we selected NO2(GT) as the target variable, representing nitrogen dioxide concentration in the air. To convert this into a binary classification task, we set a threshold of 50 µg/m³, labeling instances as 1 (high pollution) if $NO2(GT) \geq 50$ and 0 (low pollution) otherwise. This transformation allows us to evaluate classification models in distinguishing between high and low pollution levels.

    \item \textbf{Rice (Cammeo and Osmancik) Dataset} \cite{rice_(cammeo_and_osmancik)_545}:
    This dataset comprises a total of 3,810 images of rice grains from two species, Cammeo and Osmancik. The images were processed to extract morphological features that characterize the grains. Each rice grain is described by seven morphological attributes, allowing for species classification.

    \item \textbf{National Health and Nutrition Examination Survey (NHANES) Age Prediction Subset} \cite{national_health_and_nutrition_health_survey_2013-2014_(nhanes)_age_prediction_subset_887}:
    This dataset is a subset of the NHANES 2013-2014 data, obtained from the \href{https://wwwn.cdc.gov/nchs/nhanes/search/DataPage.aspx?Component=Questionnaire&CycleBeginYear=2013}{Centers for Disease Control and Prevention (CDC)}. It includes demographic and health-related features used to classify individuals into two age groups: Adult and Senior.

    \item \textbf{Iranian Telecom Churn Dataset} \cite{iranian_churn_563}:
    This dataset was randomly collected from an Iranian telecom company’s database over a 12-month period. It consists of 3,150 customer records with 13 features, including call failures, frequency of SMS usage, number of complaints, number of distinct calls, subscription length, age group, charge amount, type of service, seconds of use, status, frequency of use, and customer value. All attributes, except for the churn label, represent aggregated data from the first nine months. The churn labels indicate whether a customer remained subscribed or churned at the end of the 12-month period, allowing for predictive modeling within a designated three-month planning gap.
    \item \textbf{HIGGS Dataset} \cite{higgs_280}: This dataset comprises 11 million instances with 28 features and is widely used to distinguish between Higgs boson signals and background noise in particle physics. For our experiment, we extracted a balanced subset of 100,000 instances, consisting of 50,000 signal samples and 50,000 background noise samples. Using our proposed algorithm, we identified the core and border points within both classes. We then progressively reduced the core points while preserving the border points, effectively compressing the dataset. To evaluate the impact of this reduction, we trained classification models on datasets of varying compression levels and measured their test accuracy, assessing the extent to which critical information can be retained while reducing redundancy.
    
\end{itemize}
More details about the data and data distribution can be found in appendix Table \ref{tab:exp2Data}

\subsection{Evaluation Metric}

Given the inherent imbalance in the datasets, we use the F1 score as the primary evaluation metric to compare the performance of the models. Aggregated F1 scores are computed across all experiments to provide a robust measure of effectiveness.

\section{Result}\label{sec:Result}

We present the findings from two key experiments that evaluate the effectiveness of our proposed methods. The first experiment demonstrates the advantages of borderline-aware oversampling, comparing its performance against traditional indiscriminate oversampling of minority class instances. The second experiment examines the impact of core-aware data reduction, highlighting the extent to which downsampling can be performed without compromising classification accuracy.

\subsection{Experiment I : Oversampling borderline data}
We calculated the average performance across all datasets and evaluated the improvement of our proposed borderline oversampling method over simple oversampling. Table~\ref{tab:improvement_bins} categorizes datasets based on the percentage improvement in F1 scores.

\begin{table}[htbp]
\centering
\caption{Count of datasets grouped by improvement range in F1 scores for borderline oversampling.}
\label{tab:improvement_bins}
\begin{tabular}{lc}
\toprule
\textbf{Improvement Range (\%)} & \textbf{Count of Datasets} \\
\midrule
$>10$\% & 10 \\
5-10\% & 15 \\
1-5\% & 20 \\
$<1$\% & 10 \\
Negative or No Improvement & 2 \\
\bottomrule
\end{tabular}

\end{table}

 The results indicate that the framework provided a significant improvement, with 10 datasets achieving more than 10\% enhancement and 15 datasets showing improvements in the 5-10\% range. Moderate gains of 1-5\% were observed in 20 datasets, while only 10 datasets had minor improvements below 1\%. Notably, only two datasets showed negative or no improvement, underscoring the robustness of the proposed method in boosting classifier performance across a wide range of datasets.

Table~\ref{tab:key_metrics} presents key performance metrics summarizing the results across all datasets. The average improvement in F1 scores was 4.26\%, with the highest gain of 13.05\% observed for the dataset “abalone-20 vs 8 9 10,” demonstrating the method’s efficacy in addressing challenging class imbalances. Conversely, the lowest improvement was -1.93\%, recorded for “page-blocks-1-3 vs 4,” indicating a rare scenario where the baseline approach marginally outperformed. Importantly, negative or no improvements were observed in only two datasets, further emphasizing the consistent superiority of the proposed approach in enhancing classification performance.

Table~\ref{tab:cumulative_wins} illustrates the cumulative wins of the borderline oversampling method compared to the baseline approach across both classifiers and datasets. The results reveal a dominant performance, with borderline oversampling achieving 318 wins across classifiers compared to just 12 wins for the baseline. Similarly, the method outperformed in 48 datasets, while the baseline was superior in only 2 datasets. A more detailed analysis for each dataset is presented in the Table \ref{tab:dataset_comparison} in appendix. These results strongly validate the effectiveness and generalizability of the proposed approach in improving classification outcomes across diverse scenarios and machine learning models.

\begin{table}[htbp]
\centering
\caption{Key metrics summarizing dataset performance for borderline oversampling.}
\label{tab:key_metrics}
\begin{tabular}{lr}
\toprule
\textbf{Metric} & \textbf{Value} \\
\midrule
Total Datasets Analyzed & 50 \\
Average Improvement (\%) & 4.26 \\
Highest Improvement (\%) & 13.05 (abalone-20 vs 8 9 10) \\
Lowest Improvement (\%) & 0 (page-blocks-1-3 vs 4) \\
Count of Negative/No Improvements & 2 \\
\bottomrule
\end{tabular}

\end{table}

\begin{table}[htbp]
\centering
\caption{Cumulative wins for borderline oversampling across classifiers and datasets.}
\label{tab:cumulative_wins}
\begin{tabular}{lcc}
\toprule
\textbf{Category }& \textbf{Borderline Wins} & \textbf{Baseline Wins} \\
\midrule
Classifiers & 318 & 12 \\
Datasets & 48 & 2 \\
\bottomrule
\end{tabular}

\end{table}

The results in Table~\ref{tab:classifier_comparison} compare the average F1 scores across all datasets and oversampling algorithms, grouped by the classifier. The results demonstrate that the proposed borderline oversampling method consistently outperforms the baseline approach, achieving an average F1 score of 0.7424 compared to 0.6998, representing a notable improvement of 4.26\%. This enhancement is observed across all classifiers, with the Decision Tree Classifier showing the highest improvement (6.59\%), followed by AdaBoost (5.88\%) and other ensemble methods like Random Forest and XGBoost, which also benefit significantly. By selectively augmenting borderline minority samples, the proposed method enriches the decision boundaries more effectively, leading to better differentiation between classes and improved generalization. The consistent gains across both simple and complex models underscore the robustness and effectiveness of this targeted augmentation strategy in addressing class imbalance.

Table~\ref{tab:classifier_count} highlights the comparative effectiveness of the borderline oversampling method versus the baseline oversampling method across different classifiers. The results demonstrate a significant advantage for borderline oversampling, with the method outperforming baseline oversampling in the vast majority of cases. For instance, classifiers like AdaBoost, Decision Tree, and Random Forest show complete dominance of the borderline approach, with counts of 48, 49, and 43 wins, respectively, and no instances of baseline oversampling performing better. Similarly, CatBoost and XGBoost also exhibit overwhelming success for the borderline approach, with only one instance each favoring baseline oversampling. While classifiers like MLP and SVM occasionally favor the baseline method (5 instances each), the borderline approach still clearly dominates with counts of 44 and 41, respectively. Overall, this table underscores the efficacy of the targeted borderline augmentation strategy, which focuses on enriching critical data points, thereby achieving superior classifier performance in most cases.

\begin{table}[htbp]
\centering
\caption{Comparison of average F1 scores and improvement across classifiers between baseline and borderline oversampling methods.}
\label{tab:classifier_comparison}
\begin{tabular}{lccc}
\toprule
\textbf{Classifier} & \textbf{Baseline F1 Score (Avg)} & \textbf{Borderline F1 Score (Avg)} & \textbf{Improvement} \\
\midrule
AdaBoost & 0.6866 & 0.7454 & 0.0588 \\
CatBoost & 0.7260 & 0.7570 & 0.0310 \\
Decision Tree Classifier & 0.6646 & 0.7304 & 0.0659 \\
MLP Classifier & 0.7000 & 0.7298 & 0.0298 \\
Random Forest Classifier & 0.7139 & 0.7534 & 0.0395 \\
SVM & 0.7028 & 0.7312 & 0.0284 \\
XGBoost & 0.7048 & 0.7497 & 0.0450 \\
\textbf{Average} & 0.6998 & 0.7424 & 0.0426 \\
\bottomrule
\end{tabular}

\end{table}

\begin{table}[htbp]
\centering
\caption{Classifier-Wise Comparison of Borderline vs. Baseline Oversampling}
\label{tab:classifier_count}
\begin{tabular}{lcc}  
\toprule
\textbf{Classifier} & \textbf{Count Border Line Higher} & \textbf{Count Base Line Higher} \\
\midrule
AdaBoost & 48 & 0 \\
CatBoost & 43 & 1 \\
Decision Tree Classifier & 49 & 0 \\
MLP Classifier & 44 & 5 \\
Random Forest Classifier & 43 & 0 \\
SVM & 41 & 5 \\
XGBoost & 50 & 1 \\
\bottomrule
\end{tabular}

\end{table}

\subsection{Experiment II: Downsampling core datapoints: Preserving data quality}

Table~\ref{tab:compression_f1} presents the impact of our core-aware downsampling method across four diverse datasets: Air Quality, Rice, NHANES, and Churn. The results demonstrate that a significant portion of the dataset can be reduced while maintaining classification performance. Across all datasets, the F1 score remains largely stable up to approximately 20-25\% compression, indicating that a substantial proportion of the data—predominantly redundant or non-informative core samples—can be removed without negatively impacting classification accuracy. Beyond 30-35\% compression, a slight decline in performance is observed, though the model retains reasonable predictive capabilities, suggesting that decision boundaries remain well-represented. However, compression beyond 40\% leads to a more noticeable drop in F1 score, particularly in the Air Quality and Rice datasets, indicating the loss of critical decision-relevant samples. The Churn dataset exhibits a more gradual degradation, likely due to its structured nature, while the NHANES dataset shows an increase in F1 score with moderate compression, suggesting that some level of redundancy may have initially hindered classification.

\begin{table}[htbp]
\centering
\caption{Average F1 Score comparison at different levels of data compression for the Air Quality, Rice, NHS, and Churn datasets. Bold values indicate that the F1 score remains stable up to around 25\% compression.}
\label{tab:compression_f1}
\begin{tabular}{lcccc}
\toprule
\textbf{Compression} & \textbf{Air Quality} & \textbf{Rice} & \textbf{NHANES} & \textbf{Churn} \\
\midrule
0\% & \textit{0.9834} & \textit{0.8996} & \textit{0.1723} & \textit{0.7979} \\
5\% & \textbf{0.9845} & \textbf{0.9023} & \textbf{0.2092} & \textbf{0.8070} \\
10\% & \textbf{0.9843} & \textbf{0.9018} & \textbf{0.2315} & \textbf{0.8070} \\
15\% & \textbf{0.9844} & 0.8991 & \textbf{0.2451} & 0.7955 \\
20\% & \textbf{0.9839} & 0.8979 & \textbf{0.3385} & 0.7905 \\
24-26\% & \textbf{0.9835} & 0.8978 & \textbf{0.2928} & \textbf{0.8015} \\
30\% & 0.9813 & 0.8918 & \textbf{0.3162} & 0.7774 \\
34-36\% & 0.9790 & 0.8881 & \textbf{0.2923} & 0.7847 \\
41\% & 0.9706 & 0.8801 & \textbf{0.2749} & 0.7539 \\
54-56\% & 0.9669 & 0.8670 & \textbf{0.2855} & 0.7355 \\
60-61\% & 0.9378 & 0.8499 & \textbf{0.3187} & 0.6869 \\
\bottomrule
\end{tabular}

\end{table}

\subsubsection{Big data result}

The effectiveness of machine learning models, particularly in imbalanced classification tasks, is often hindered by the failure to differentiate between critical instances near the decision boundary and redundant samples concentrated in the core of the data distribution. In this paper, we propose a method to systematically identify and differentiate between these two types of data, improving data quality by focusing on critical decision points while reducing redundancy. Through extensive experiments on multiple benchmark datasets, we demonstrate that oversampling the borderline data improves classification performance, achieving an average increase of 4.26\% in F1 score, while our core-aware reduction method enables up to 25\% data compression without performance degradation as shown in Figure~\ref{tab:classifier_compression}. Furthermore, we apply our method to the large-scale and balanced Higgs dataset, where we achieve up to 70\% data compression with minimal accuracy loss, highlighting its scalability for big data scenarios. Beyond imbalanced classification, our method has broader implications for efficient model training, particularly in computationally expensive domains such as Large Language Model (LLM) training. By prioritizing high-quality, decision-relevant data, our approach can be extended to text, multimodal, and self-supervised learning scenarios, offering a pathway to faster convergence, improved generalization, and significant computational savings. This work paves the way for future research in data-efficient learning, where intelligent sampling replaces brute-force expansion, driving the next generation of AI advancements.

\begin{table}[htbp]
\centering
\caption{Test accuracy across different classifiers under varying levels of dataset compression. The results show how classification performance is maintained as we progressively remove redundant core points while preserving border points.}
\label{tab:classifier_compression}
\begin{tabular}{p{1.5cm}p{1.5cm}p{1.5cm}p{1.5cm}p{1.5cm}p{1.5cm}}

\toprule
\textbf{Compr- ession (\%)} & \textbf{Random Forest} & \textbf{MLP} & \textbf{SVM} & \textbf{Decision Tree} & \textbf{Ada Boost} \\
\midrule
0  & \textit{0.709} & \textit{0.662} & \textit{0.666} & \textit{0.624} & \textit{0.700} \\
20 & \textbf{0.709} & \textbf{0.704} & \textbf{0.668} & \textbf{0.634} & 0.691 \\
25 & \textbf{0.71} & \textbf{0.69} & \textbf{0.671} & \textbf{0.63} & 0.69 \\
28 & \textbf{0.71} & \textbf{0.712} & \textbf{0.67} & \textbf{0.633} & 0.689 \\
30 & \textbf{0.712} & \textbf{0.708} & \textbf{0.671} & \textbf{0.636} & 0.689 \\
35 & \textbf{0.71} & \textbf{0.71} & \textbf{0.671} & \textbf{0.631} & 0.69 \\
38 & \textbf{0.709} & \textbf{0.689} & \textbf{0.672} & \textbf{0.635} & 0.69 \\
40 & 0.705 & \textbf{0.697} & \textbf{0.671} & \textbf{0.63} & 0.692 \\
45 & 0.707 & \textbf{0.679} & \textbf{0.668} & \textbf{0.635} & 0.693 \\
48 & \textbf{0.712} & \textbf{0.704} & \textbf{0.668} & \textbf{0.638} & 0.695 \\
50 & 0.708 & \textbf{0.7} & \textbf{0.668} & \textbf{0.629} & 0.694 \\
55 & \textbf{0.71} & \textbf{0.681} & \textbf{0.667} & \textbf{0.632} & \textbf{0.7} \\
58 & \textbf{0.709} & \textbf{0.662} & \textbf{0.666} & \textbf{0.624} & \textbf{0.7} \\
60 & \textbf{0.71} & \textbf{0.684} & \textbf{0.666} & \textbf{0.633} & 0.698 \\
65 & \textbf{0.709} & \textbf{0.688} & \textbf{0.667} & \textbf{0.637} & 0.699 \\
67 & \textbf{0.71} & \textbf{0.683} & 0.665 & \textbf{0.638} & 0.697 \\
70 & \textbf{0.709} & \textbf{0.687} & 0.661 & \textbf{0.637} & 0.697 \\
75 & \textbf{0.71} & \textbf{0.682} & 0.654 & \textbf{0.633} & \textbf{0.7} \\
77 & 0.708 & \textbf{0.684} & 0.653 & \textbf{0.639} & \textbf{0.701} \\
80 & \textbf{0.709} & \textbf{0.681} & 0.65 & \textbf{0.643} & \textbf{0.701} \\
85 & 0.703 & \textbf{0.669} & 0.649 & \textbf{0.639} & 0.699 \\
88 & 0.701 & \textbf{0.665} & 0.643 & \textbf{0.632} & 0.695 \\
90 & 0.696 & 0.653 & 0.638 & \textbf{0.632} & 0.698 \\
95 & 0.693 & 0.658 & 0.63 & \textbf{0.628} & 0.688 \\
98 & 0.661 & 0.562 & 0.587 & 0.617 & 0.674 \\
99 & 0.658 & 0.615 & 0.6 & 0.618 & 0.671 \\
\bottomrule
\end{tabular}
\end{table}

Overall, these findings emphasize the efficacy of our method in data-efficient learning, allowing for dataset reduction without sacrificing model performance. This has direct implications for improving computational efficiency in machine learning workflows, particularly in resource-constrained environments where optimizing data volume is essential.

\section{Conclusion and Future Work}\label{sec:Conc}

In this study, we proposed a novel borderline oversampling approach that selectively augments critical minority class instances while preserving the decision boundary, as well as a data reduction technique that prunes redundant core samples while maintaining classification performance. Our experimental results demonstrate that the borderline oversampling method significantly improves classifier performance across various datasets, with an average F1 score improvement of 4.26\%. This targeted augmentation approach effectively enriches the minority class distribution, leading to better generalization and superior classification outcomes across different machine learning models.

In addition, our data reduction strategy confirms that substantial portions of the dataset—particularly redundant core samples—can be removed without negatively impacting classifier performance. Our findings show that up to 25\% of the data can be pruned while maintaining an equivalent or even slightly improved F1 score. Beyond this threshold, the model still retains high performance, with only marginal degradation observed up to 35\% compression, before a significant decline is noted at higher compression levels. These results highlight the effectiveness of strategic data selection in enhancing computational efficiency while preserving decision boundaries.

More broadly, our work underscores the importance of data quality over quantity in machine learning. In traditional imbalanced classification settings, blindly oversampling all minority instances or retaining all majority instances often leads to unnecessary redundancy and increased computational cost. Our borderline-aware augmentation and core-aware pruning techniques demonstrate that classifiers benefit more from well-structured, high-quality data rather than simply increasing dataset size.


Our method presents a promising generalized framework for identifying high-quality data, which has far-reaching implications beyond class imbalance problems. One of the most compelling extensions of this work is its potential application in Large Language Model (LLM) training. Given the high cost and resource intensity of training LLMs, our approach could be adapted to text data by identifying and prioritizing high-quality and informative text samples while filtering out redundant or low-impact data. By applying borderline-aware text augmentation and core-aware text pruning, LLMs could achieve faster convergence and improved generalization with reduced training costs.

Another key direction involves adapting our approach to multimodal data, where both structured (tabular) and unstructured (image, text) data require selective augmentation and pruning. Exploring the impact of borderline-aware and core-aware sampling in self-supervised learning, reinforcement learning, and active learning scenarios could further solidify the effectiveness of data quality-driven training strategies.

In conclusion, our findings demonstrate that intelligent data selection, rather than brute-force dataset expansion, is a key enabler of efficient and effective machine learning models. As training data continues to scale in modern AI, our method offers a compelling solution for optimizing training efficiency while maintaining or even enhancing model performance.

\section{Appendix}\label{sec:appendix}

\footnotesize
\begin{longtblr}[
  caption = {Dataset Characteristics: Number of Rows, Columns, and Data Distribution for Each Dataset Used in the Experiments},
  label = {tab:dataset_characteristics},
  entry = {Spanning Table Example},
]{
  colspec = {|X[1.65]|X[1]|X[1.4]|X[2.7]|}, 
  rowhead = 1, 
  hlines,
  vlines,
}
\hline
\textbf{Dataset}        & \textbf{\# Rows} & \textbf{\# Columns} & \textbf{Data Distribution (Class 0:Class 1)} \\
\hline
abalone19 & 4174 & 9 & 0: 4142 1: 32 \\
abalone9 18 & 731 & 9 & 0: 689 1: 42 \\
abalone-17 vs 7-8-9-10 & 2338 & 9 & 0: 2280 1: 58 \\
abalone-19 vs 10-11-12-13 & 1622 & 9 & 0: 1590 1: 32 \\
abalone-20 vs 8 9 10 & 1916 & 9 & 0: 1890 1: 26 \\
abalone-22 vs 8 & 581 & 9 & 0: 567 1: 14 \\
abalone-3 vs 11 & 502 & 9 & 0: 487 1: 15 \\
ADA & 4147 & 48 & 0: 3118 1: 1029 \\
appendi citis & 106 & 7 & 0: 85 1: 21 \\
bupa & 345 & 6 & 0: 200 1: 145 \\
cleveland-0 vs 4 & 177 & 13 & 0: 164 1: 13 \\
CM1 & 498 & 21 & 0: 449 1: 49 \\
dermato logy-6 & 358 & 34 & 0: 338 1: 20 \\
ecoli1 & 336 & 7 & 0: 259 1: 77 \\
ecoli2 & 336 & 7 & 0: 284 1: 52 \\
ecoli3 & 336 & 7 & 0: 301 1: 35 \\
ecoli4 & 336 & 7 & 0: 316 1: 20 \\
ecoli 0 1 3 7 vs 2 6 & 281 & 7 & 0: 274 1: 7 \\
ecoli 0 1 4 6 vs 5 & 280 & 6 & 0: 260 1: 20 \\
ecoli 0 1 4 7 vs 2 3 5 6 & 336 & 7 & 0: 307 1: 29 \\
ecoli 0 1 4 7 vs 5 6 & 332 & 6 & 0: 307 1: 25 \\
ecoli 0 1 vs 2 3 5 & 244 & 7 & 0: 220 1: 24 \\
ecoli 0 1 vs 5 & 240 & 6 & 0: 220 1: 20 \\
ecoli 0 2 3 4 vs 5 & 202 & 7 & 0: 182 1: 20 \\
ecoli 0 2 6 7 vs 3 5 & 224 & 7 & 0: 202 1: 22 \\
ecoli 0 3 4 6 vs 5 & 205 & 7 & 0: 185 1: 20 \\
ecoli 0 3 4 7 vs 5 6 & 257 & 7 & 0: 232 1: 25 \\
ecoli 0 4 6 vs 5 & 203 & 6 & 0: 183 1: 20 \\
ecoli 0 6 7 vs 3 5 & 222 & 7 & 0: 200 1: 22 \\
ecoli 0 6 7 vs 5 & 220 & 6 & 0: 200 1: 20 \\
ecoli 0 vs 1 & 220 & 7 & 0: 143 1: 77 \\
flare-F & 1066 & 30 & 0: 1023 1: 43 \\
glass0 & 214 & 9 & 0: 144 1: 70 \\
glass1 & 214 & 9 & 0: 138 1: 76 \\
glass2 & 214 & 9 & 0: 197 1: 17 \\
glass4 & 214 & 9 & 0: 201 1: 13 \\
glass5 & 214 & 9 & 0: 205 1: 9 \\
glass 0 1 5 vs 2 & 172 & 9 & 0: 155 1: 17 \\
glass 0 1 6 vs 2 & 192 & 9 & 0: 175 1: 17 \\
glass 0 1 6 vs 5 & 184 & 9 & 0: 175 1: 9 \\
glass 0 4 vs 5 & 92 & 9 & 0: 83 1: 9 \\
glass 0 6 vs 5 & 108 & 9 & 0: 99 1: 9 \\
hepatitis & 155 & 19 & 0: 123 1: 32 \\
iris0 & 150 & 4 & 0: 100 1: 50 \\
kddcup-land vs portsweep & 1061 & 68 & 0: 1040 1: 21 \\
led7digit-0-2-4-6-7-8-9 vs 1 & 443 & 7 & 0: 406 1: 37 \\
monk-2 & 432 & 6 & 0: 228 1: 204 \\
page-blocks-1-3 vs 4 & 472 & 10 & 0: 444 1: 28 \\
PC1 & 1109 & 21 & 0: 1032 1: 77 \\
saheart & 462 & 9 & 0: 302 1: 160 \\
wdbc & 569 & 30 & 0: 357 1: 212 \\
yeast-0-2-5-6 vs 3-7-8-9 & 1004 & 8 & 0: 905 1: 99 \\
yeast-0-2-5-7-9 vs 3-6-8 & 1004 & 8 & 0: 905 1: 99 \\
yeast-0-3-5-9 vs 7-8 & 506 & 8 & 0: 456 1: 50 \\
\hline
\end{longtblr}

\begin{longtblr}[
  caption = {Comparison of Average F1 Scores and Improvement for Different Datasets Between Baseline and Borderline Oversampling Methods},
  label = {tab:dataset_comparison},
  entry = {Spanning Table Example},
]{
  colspec = {|X[1.75]|X[1.5]|X[1.5]|X[1.25]|}, 
  rowhead = 1, 
  hlines,
  vlines
}
\hline
\textbf{Dataset} & \textbf{Base line F1 Score (Avg)} & \textbf{Border line F1 Score (Avg)} & \textbf{Improve- ment} \\
\hline
abalone-20 vs 8 9 10 & 0.3713 & 0.5018 & 0.1305 \\
abalone-22 vs 8 & 0.6167 & 0.7391 & 0.1224 \\
glass5 & 0.7314 & 0.8467 & 0.1152 \\
glass4 & 0.7639 & 0.8544 & 0.0905 \\
ecoli 0 1 4 6 vs 5 & 0.7873 & 0.8646 & 0.0773 \\
abalone9 18 & 0.4428 & 0.5181 & 0.0752 \\
ecoli 0 6 7 vs 3 5 & 0.7733 & 0.8482 & 0.0749 \\
ecoli 0 1 vs 2 3 5 & 0.7891 & 0.8625 & 0.0735 \\
abalone-17 vs 7-8-9-10 & 0.3543 & 0.4214 & 0.0671 \\
hepatitis & 0.6039 & 0.6707 & 0.0667 \\
ecoli 0 1 3 7 vs 2 6 & 0.6467 & 0.7133 & 0.0667 \\
abalone19 & 0.0794 & 0.1394 & 0.0600 \\
cleveland-0 vs 4 & 0.6956 & 0.7543 & 0.0587 \\
ecoli 0 3 4 6 vs 5 & 0.8210 & 0.8779 & 0.0568 \\
ecoli 0 2 6 7 vs 3 5 & 0.7696 & 0.8255 & 0.0559 \\
flare-F & 0.2460 & 0.3016 & 0.0556 \\
CM1 & 0.2853 & 0.3405 & 0.0552 \\
ecoli3 & 0.6657 & 0.7203 & 0.0547 \\
appendicitis & 0.6397 & 0.6935 & 0.0538 \\
glass2 & 0.4553 & 0.5068 & 0.0516 \\
ecoli2 & 0.8202 & 0.8706 & 0.0505 \\
PC1 & 0.3850 & 0.4320 & 0.0470 \\
yeast-0-3-5-9 vs 7-8 & 0.4407 & 0.4861 & 0.0454 \\
ecoli 0 1 vs 5 & 0.8389 & 0.8840 & 0.0451 \\
glass 0 1 6 vs 5 & 0.8400 & 0.8829 & 0.0429 \\
Average & 0.6998 & 0.7424 & 0.0426 \\
ecoli4 & 0.8215 & 0.8632 & 0.0417 \\
ecoli 0 1 4 7 vs 2 3 5 6 & 0.7822 & 0.8232 & 0.0410 \\
abalone-19 vs 10-11-12-13 & 0.1382 & 0.1786 & 0.0404 \\
ecoli 0 4 6 vs 5 & 0.8345 & 0.8748 & 0.0404 \\
saheart & 0.5587 & 0.5963 & 0.0376 \\
ecoli 0 2 3 4 vs 5 & 0.8538 & 0.8912 & 0.0374 \\
glass 0 1 5 vs 2 & 0.4341 & 0.4702 & 0.0362 \\
glass0 & 0.7717 & 0.8051 & 0.0334 \\
ecoli 0 1 4 7 vs 5 6 & 0.8267 & 0.8598 & 0.0331 \\
yeast-0-2-5-6 vs 3-7-8-9 & 0.5923 & 0.6249 & 0.0325 \\
ecoli 0 3 4 7 vs 5 6 & 0.8247 & 0.8569 & 0.0322 \\
ecoli1 & 0.8008 & 0.8298 & 0.0290 \\
glass1 & 0.7423 & 0.7709 & 0.0286 \\
ecoli 0 6 7 vs 5 & 0.8245 & 0.8503 & 0.0258 \\
glass 0 6 vs 5 & 0.9552 & 0.9790 & 0.0238 \\
yeast-0-2-5-7-9 vs 3-6-8 & 0.8166 & 0.8388 & 0.0222 \\
bupa & 0.6613 & 0.6822 & 0.0209 \\
glass 0 4 vs 5 & 0.9543 & 0.9733 & 0.0190 \\
abalone-3 vs 11 & 0.9837 & 0.9959 & 0.0122 \\
led7digit-0-2-4-6-7-8-9 vs 1 & 0.8074 & 0.8173 & 0.0099 \\
dermatology-6 & 0.9873 & 0.9968 & 0.0095 \\
wdbc & 0.9551 & 0.9639 & 0.0088 \\
ADA & 0.6570 & 0.6654 & 0.0084 \\
ecoli 0 vs 1 & 0.9834 & 0.9853 & 0.0019 \\
monk-2 & 0.9939 & 0.9956 & 0.0017 \\
iris0 & 0.9986 & 1.0000 & 0.0014 \\
kddcup-land vs portsweep & 1.0000 & 1.0000 & 0.0000 \\
glass 0 1 6 vs 2 & 0.4016 & 0.4008 & -0.0008 \\
page-blocks-1-3 vs 4 & 0.9649 & 0.9456 & -0.0193 \\
\hline
\end{longtblr}

\normalsize %
\begin{table}[htbp]
\centering
\caption{Summary of datasets used in Experiment II}
\label{tab:exp2Data}
\begin{tabular}{llll}
\toprule
\textbf{Dataset} & \textbf{\# Rows} & \textbf{\# Columns} &\textbf{Data Distribution} \\
\midrule
Air Quality & 9357 & 12 & 0: 2220, 1: 7137  \\
Rice & 3810 & 7 & 0: 2180, 1: 1630  \\
NHANES & 2278 & 7 & 0: 1914, 1: 364 \\
Churn & 3150 & 13 & 0: 2655, 1: 495  \\
Higgs Dataset & 100000 & 21 & 0: 50000, 1: 50000  \\
\bottomrule
\end{tabular}

\end{table}

\section{Acknowledgements}
We extend our heartfelt appreciation to Qatar National Library for their invaluable support in covering the publication charge.

\bibliography{main}

\end{document}